\documentclass[journal,twoside,web]{ieeecolor}
\usepackage{tmi}
\usepackage{cite}
\usepackage{amsmath,amssymb,amsfonts}
\usepackage{algorithmic}
\usepackage{graphicx}
\usepackage{textcomp}

\usepackage{multirow}
\usepackage{subfig}
\usepackage{tabularx}
\newcolumntype{Y}{>{\centering\arraybackslash}X}
\usepackage[export]{adjustbox}

\newcommand{\ie}{\emph{i.e.,}~}
\newcommand{\eg}{\emph{e.g.,}~}

\usepackage{arydshln}
  
\usepackage{times}
\usepackage{epsfig}
\usepackage{graphicx}
\usepackage{amsmath}
\usepackage{amssymb}
\usepackage{breqn}
\usepackage{algorithm,algorithmic}

\usepackage{makecell}

\usepackage{multirow}
\usepackage{subfig}
\usepackage{tabularx}
\newcolumntype{Y}{>{\centering\arraybackslash}X}
\usepackage[export]{adjustbox}

\usepackage{amsmath}
\usepackage{amsfonts}

\DeclareMathOperator{\D}{\mathcal{D}}
\DeclareMathOperator{\X}{\mathcal{X}}
\DeclareMathOperator{\K}{\mathcal{K}}

\def\BibTeX{{\rm B\kern-.05em{\sc i\kern-.025em b}\kern-.08em
    T\kern-.1667em\lower.7ex\hbox{E}\kern-.125emX}}
\begin{document}
\title{U-LanD: Uncertainty-Driven Video Landmark Detection}

\author{Mohammad H. Jafari, Christina Luong, Michael Tsang, Ang Nan Gu, Nathan Van Woudenberg, Robert Rohling, Teresa Tsang, Purang Abolmaesumi

\thanks{This work was supported in part by the Natural
Sciences and Engineering Research Council (NSERC) of Canada and in part by the Canadian
Institutes of Health Research (CIHR). T. Tsang is the Director of the Vancouver General Hospital and University of British Columbia Echocardiography Laboratories, and Principal Investigator of the CIHR-NSERC grant supporting this work. P. Abolmaesumi is Co-Principal Investigator of the grant supporting this work. T. Tsang and P. Abolmaesumi contributed equally (joint senior authors).}
\thanks{M. Jafari, A. Gu, N. Woudenberg, and P. Abolmaesumi are with the Department of Electrical and Computer Engineering, The University of British Columbia, Vancouver, BC V6T 1Z4, Canada (e-mail: purang@ece.ubc.ca).}
\thanks{R. Rohling is with the Department of Electrical and Computer Engineering, The University of British Columbia, Vancouver, BC V6T 1Z4,Canada, and also with the Department of Mechanical Engineering, The University of British Columbia, Vancouver, BC V6T 1Z4, Canada.}
\thanks{C. Luong, M. Tsang, and T. Tsang are with the Vancouver General Hospital, Echocardiography Laboratory, Division of Cardiology, Department of Medicine, The University of British Columbia, Vancouver, BC V5Z 1M9, Canada (e-mail: t.tsang@ubc.ca).}}



\maketitle

\begin{abstract}
This paper presents U-LanD, a framework for joint detection of key frames and landmarks in videos. We tackle a specifically challenging problem, where training labels are noisy and highly sparse. U-LanD builds upon a pivotal observation: a deep Bayesian landmark detector solely trained on key video frames, has significantly lower predictive uncertainty on those frames vs. other frames in videos. We use this observation as an unsupervised signal to automatically recognize key frames on which we detect landmarks. As a test-bed for our framework, we use ultrasound imaging videos of the heart, where sparse and noisy clinical labels are only available for a single frame in each video. Using data from 4,493 patients, we demonstrate that U-LanD can exceedingly outperform the state-of-the-art non-Bayesian counterpart by a noticeable absolute margin of 42\% in $R^2$ score, with almost no overhead imposed on the model size. Our approach is generic and can be potentially applied to other challenging data with noisy and sparse training labels. 

\end{abstract}


\section{Introduction}
\label{sec:introduction}

\begin{figure}[t!]
    \centering
    \includegraphics[width=.98\columnwidth]{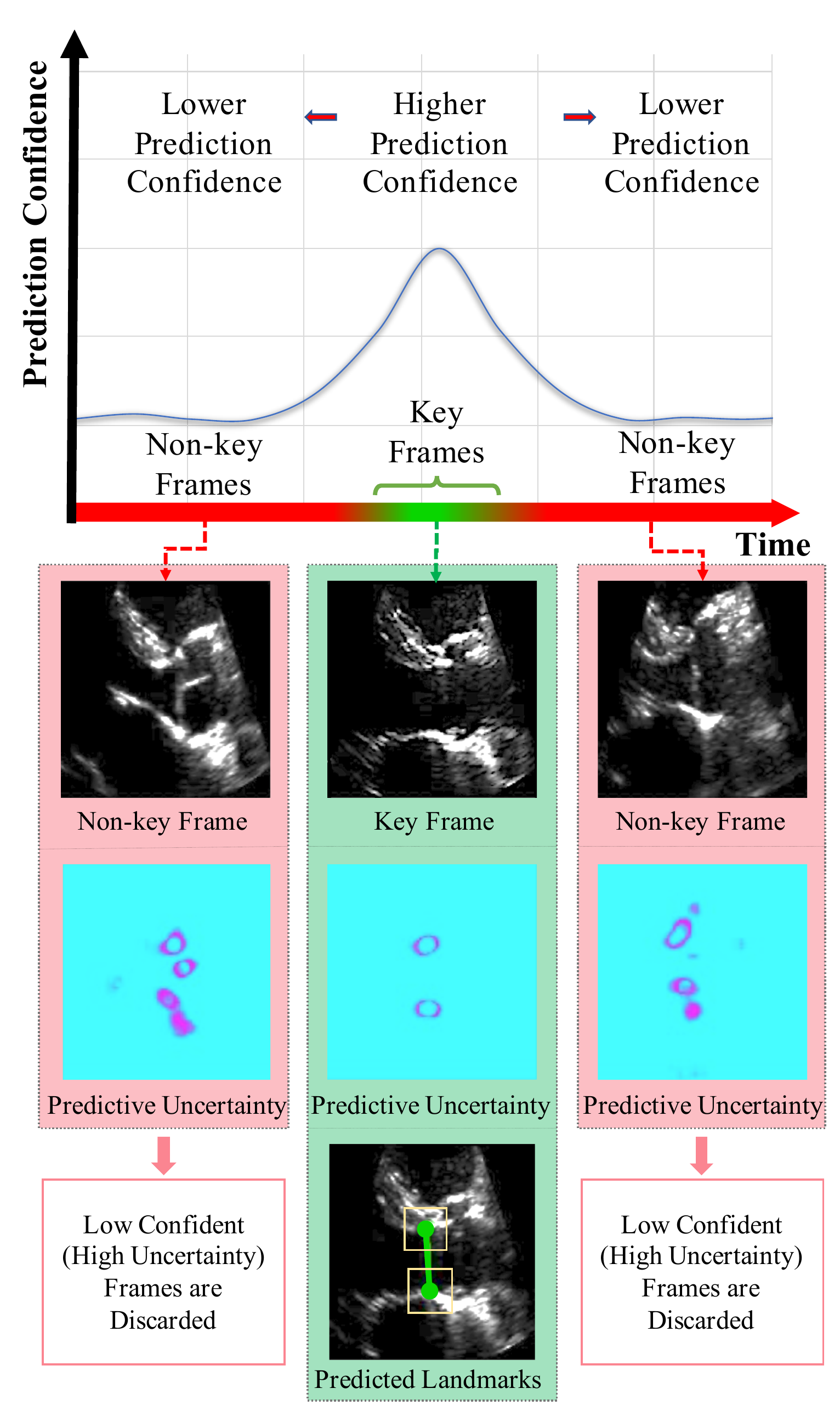}
    \caption{\textbf{Key Observation:} A Bayesian landmark detector has a significantly different predictive uncertainty on key frames vs. non-key frames. Our proposed U-LanD framework exploits this observation to jointly identify key frames and predict landmarks in videos. We demonstrate the efficacy of U-LanD for the very challenging problem of analyzing ultrasound videos of the heart, where only single-frame, noisy labels are available for each training sample.  
     }
    \label{fig:overview}
\end{figure}

Accurate detection of the reference points (landmarks) coordinates in images and videos is a fundamental step in many vision tasks, such as face analysis~\cite{chandran2020attention,chen2019deep,ma2020deep,nartey2019semi,sun2019fab, kumar2020luvli}, pose estimation~\cite{yu2019layout,zhang2020distribution,wei2016convolutional}, and medical imaging~\cite{bhalodia2020self,noothout2020deep,yao2020miss}. 

The video landmark detection involves additional complexities of recognizing key frames used to detect the landmarks. The key frames denote moments in time when a targeted event happens, and an object of interest has the highest visibility in the video. During data labelling routines, experts usually only label the landmark points at the selected key frames. As a result, the available training video datasets suffer from two limitations: 1) videos are sparsely labelled, \ie a small portion of frames in each video have ground-truth landmark labels; and 2) the labelled frames are extensively biased towards specific points in time, \ie only key frames in each training video are labelled. 
Previous work mainly divides the problem of video object detection on key frames into sub-problems of key frame recognition~\cite{griffin2019bubblenets,kulhare2016key,mishra2020real,wu2019adaframe,yan2018deep} and object detection. They propose techniques such as self-supervised learning~\cite{aakur2019perceptual}, semi-supervised learning~\cite{chen2020semi}, label propagation~\cite{zhu2019improving}, registration~\cite{dong2020supervision}, and temporal cycle-consistency~\cite{wang2019learning}. However, the inherent predictive uncertainties in video landmark measurements are generally ignored in the existing literature. 

In this paper, we present \textit{U-LanD}, a novel uncertainty-driven video landmark detection framework. U-LanD is specifically designed to consider the bias in sparsely annotated video data and automatically detect the location of landmarks on the key frames of the video. U-LanD is based on a core observation: a deep Bayesian encoder-decoder network only trained on labelled key frames, shows significantly higher prediction confidence (lower uncertainty) in test time when deployed on key frames vs. non-key frames, as illustrated in Fig~\ref{fig:overview}.
U-LanD exploits the test-time  
predictive uncertainty as an unsupervised guiding signature to detect landmark points on key frames. U-LanD considers several aspects of predictive uncertainty in landmark detection, namely i) contextual quality control based on 
landmark heatmaps; ii) pixel-level epistemic uncertainty with variational dropout; and iii) pixel-level heteroscedastic aleatoric uncertainty by placing a Gaussian distribution over the observed labels.


We demonstrate the efficacy of U-LanD on a challenging dataset with sparsely annotated data, namely echocardiography (echo, heart ultrasound).  
Echo videos, aka echo cine series, have a notoriously noisy nature (sample frames can be seen in Fig.~\ref{fig:overview}), which increases the complication for automated analysis. 
We tackle the task of left ventricular outflow tract (LVOT) landmark detection~\cite{mitchell2019guidelines}, as a problem that suffers from an extreme limit of label sparsity; in LVOT training videos, only \textit{one} frame throughout the entire span of frames has the ground-truth landmark label. In each training video, the expert clinician has annotated locations of two points (coordinates of points used to measure the LVOT diameter), only for one frame (a frame around the mid-systole phase of the heart), where the targeted object, \ie the aorta, has the highest visibility. For our experiments, we gather a large-scale echo dataset of 4,493 patients, demonstrating that U-LanD can significantly improve the results of state-of-the-art non-Bayesian counterparts. Also, while U-LanD is fully automatic, it can surpass the results of semi-automatic key frame landmark detection involving expert supervision.

\section{Related Works}
\label{sec:related}
\subsection{Image Landmark Detection}
Deep models are the de-facto state-of-the-art for image landmark detection, which can be categorized into three broad groups: i) Regression networks~\cite{Lv_2017_CVPR, zhu2015face}, which use Convolutional Neural Networks (CNNs) to directly regress coordinate values of the landmark points; ii) Reinforcement Learning (RL) approaches~\cite{alansary2019evaluating, vlontzos2019multiple, zhao2020deep}, where the RL agents learn to find an optimal path to the locations of the landmarks by interacting with the surrounding environment in the image space; and iii) Fully Convolutional Networks (FCN)~\cite{gilbert2019automated,Liu_2019_CVPR,merget2018robust, wei2016convolutional}, which tackle landmark detection as semantic segmentation by predicting heatmaps (or blobs) denoting the locations of landmarks. Various iterative or cascaded combinations of the above approaches are used in the literature~\cite{valle2018deeply,wang2019adaptive}. The FCN approach provides advantages such as translation invariance, independence from image size, and interpretability of the predicted heatmaps. The FCN networks~\cite{long2015fully} have benefited from many improvements in recent years through successor architectures, such as the U-Net~\cite{unet}, Deeplab family~\cite{chen2018encoder}, and Efficientdet~\cite{tan2020efficientdet}. 
In this work, we build U-LanD on top of a modified U-Net model, while the method has no specific assumptions regarding the choice of the backbone landmark detection network.

\subsection{Video Landmark Tracking}
In video landmark tracking, the goal is to label all the frames of the video and provide temporally-consistent predictions through time. Variations of the above image landmark detectors are extended to temporal dimensions using 3D/(2+1)D models~\cite{tran2015learning,tran2018closer}, attention modules~\cite{yu2020deformable},  transformers~\cite{girdhar2019video}, and models based on optical flow tracking~\cite{sun2019fab}. 
However, our task requires only to detect landmarks on the key frames of the video and discard the measurements on non-key frames; therefore, the stabilized tracking through time, followed by the above researches, is not the goal here.  

\subsection{Label Propagation}
The semi-supervised and self-supervised learning techniques~\cite{aakur2019perceptual,bhalodia2020self,chen2020semi} are used to improve training when data is partly labelled, \eg only few frames of training videos have ground-truth annotation. Methods such as registration~\cite{dong2020supervision} and temporal correspondence by cycle-consistency~\cite{li2019joint, peng2020chained,wang2019learning} are used to propagate labels to unlabelled frames, providing weakly supervised annotations to reduce the bias in training~\cite{chen2020learning, nishimura2020weakly}. Our work is different from this line of research, where U-LanD leverages the existing training bias, captured via predictive uncertainty, as a signature to recognize the key frames. 


\subsection{Prediction Quality Control}
Safe deployment of deep learning methods requires the model not to report a result with low confidence (high uncertainty), \ie raise an alarm that the model \textit{does not know} the prediction.  Works of~\cite{devries2018leveraging,jungo2018uncertainty, kwon2018uncertainty} leverage the predictive uncertainty to detect 
failed predictions. Also, non-uncertainty based approaches include scoring the validity of the prediction by comparing it to the distribution of ground-truth labels, via approaches such as atlas registration~\cite{robinson2017automatic}, 
generative adversarial networks~\cite{schlegl2017unsupervised}, and variational autoencoders~\cite{liu2019alarm}. However, in U-LanD, we track the prediction confidence through time and use it to detect landmarks on the key frames of the video. 
To this end, U-LanD measures pixel-level epistemic and aleatoric uncertainty, and a task-specific contextual validity check based on the predicted landmark heatmaps.

\begin{figure*}[t!]
    \centering
    \includegraphics[width=0.98\textwidth]{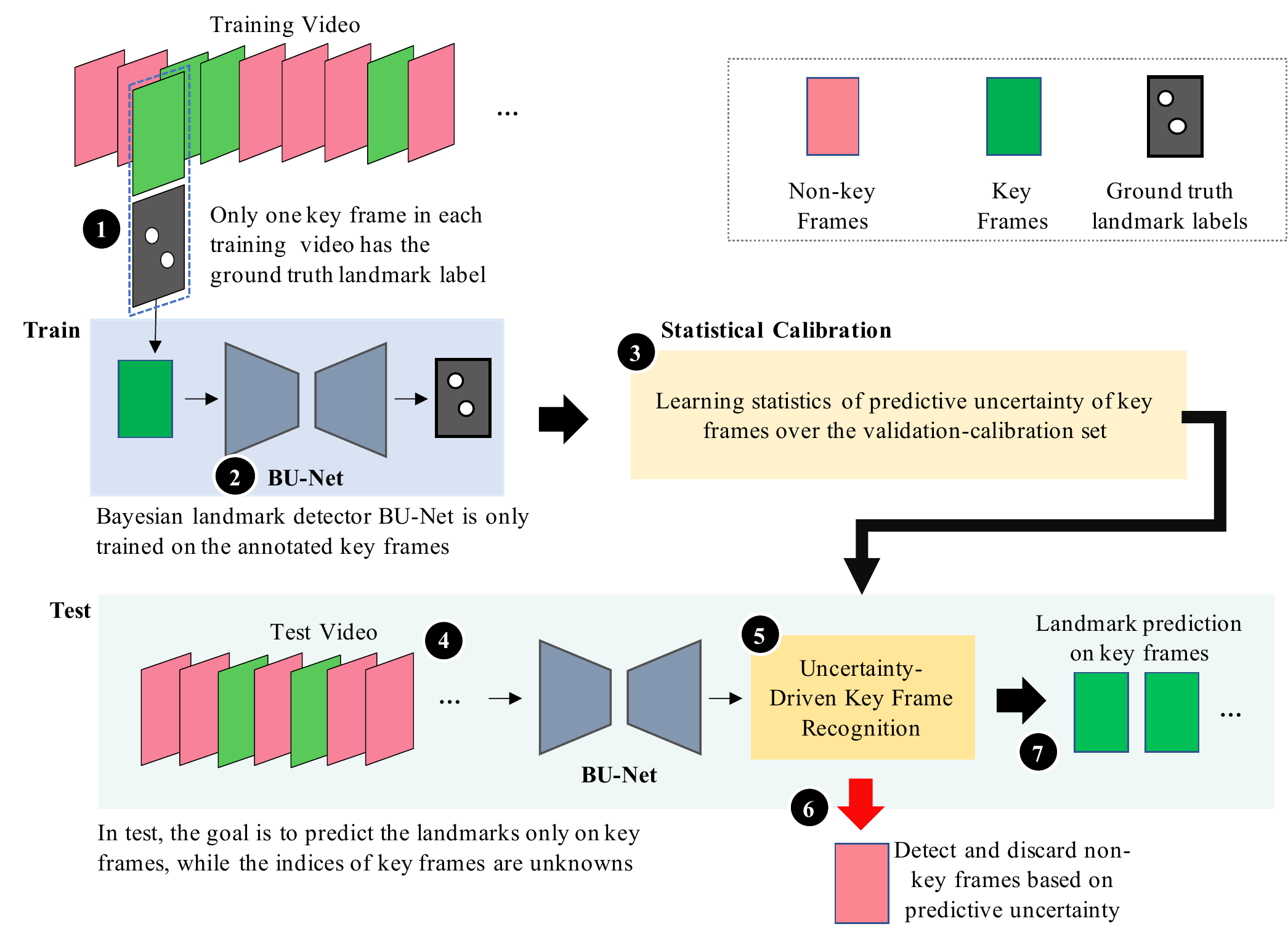}
    \caption{A block diagram of the proposed U-LanD framework, including training, statistical calibration, and test phases. U-LanD automatically predicts landmarks on key frames of videos, where the training videos are only labelled on one frame. We propose to leverage the predictive uncertainty of a Bayesian landmark detector (BU-Net) to recognize key frames vs. non-key frames through time.}
    \label{fig:block_diagram}
\end{figure*}

\section{Contributions}
In this work, we preset U-LanD, a deep Bayesian framework for landmark detection on key frames of the videos. 
To summarize, the major contributions of the paper are:
\begin{itemize}
    \item We formulate a Bayesian method for video landmark detection using sparsely labelled training data; only one key frame of each training video has the ground-truth label.
    \item We exploit the inter-frame variations of predictive uncertainty as an unsupervised signal to recognize key frames. This  enables a simple 2D landmark detector to achieve superior results for video landmark detection.
    
    \item We demonstrate U-LanD in a challenging task of echo key frame landmark detection. We gather a large-scale echo video dataset of 4,493 patients for our experiments, showing U-LanD can improve the state-of-the-art by a significant absolute margin of 42\% in $R^2$ score, and surprisingly surpass semi-automatic key frame landmark detection by expert human supervision.
\end{itemize}


\section{Proposed Method: U-LanD}
\subsection{Problem Formulation}
\label{Sec:formalized}
The goal is to automatically detect landmark locations on key frames of the video, while in training data only one key frame of each video has the ground-truth landmark, as shown in 
{\centering\includegraphics[width=0.025\textwidth,valign=c]{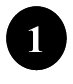}}
in Fig.~\ref{fig:block_diagram}. The inputs at test time are variable-length videos, with no human supervision regarding location of landmarks or indices of the key frames, as shown in {\centering\includegraphics[width=0.025\textwidth,valign=c]{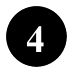}} in Fig.~\ref{fig:block_diagram}. To define the task formally, let $\D = \{(f^{k_i}_i,y^{k_i}_i)\}_{i=1}^{N}$ indicate the $N$ training videos, where pairs of $f^{k_i}_i$ and $y^{k_i}_i$ are the frame and ground-truth labels (coordinates of $\tau$ landmarks) on the key frame $k_i$ of the $i^{th}$ training video. 
Each video may have one or more key frames, represented by ${\K_i}$ set, but only one is labelled, \ie $k_i \in {\K_i}$, $|\K_i| \geq 1$. Sample of test data is denoted by $\hat x_i = {\hat f_{i}}^{ \ 1:P_i}$, which is the $i^{th}$ test video with $P_i$ frames; test videos can have variable lengths. The U-LanD framework $U(x)$ aims to obtain the mapping $U({\hat f_{i}}^{ \ 1:P_i}) \xrightarrow[]{} \hat y^{k_i}_i$, \ie the $U(x)$ predicts the key frame landmark measurements. The key frames ($k$) in videos can occur at different indices which are unknowns, but all the key frames have specific common features, such as denoting specific events, when a targeted object has a high visibility. 

\subsection{Solution Overview}
The block diagram of the train, calibration, and test phases of the U-LanD framework are shown in Figure~\ref{fig:block_diagram}. In U-LanD, a Bayesian 2D image landmark detector, namely \textit{BU-Net}, is solely trained on the labelled key frame in $\D$, as is shown in {\centering\includegraphics[width=0.025\textwidth,valign=c]{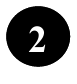}} in Fig.~\ref{fig:block_diagram}. The BU-Net is a deep Bayesian encoder-decoder network, formulated in Section~\ref{sec:backbone}, predicting both the heatmaps ($\hat y$) that show the locations of landmarks, as well as the predictive uncertainty maps~(epistemic $\hat u_{epi}$ and aleatoric $\hat u_{alea}$) that show pixel classification uncertainties. 

In training, BU-Net has only been exposed to key frames, therefore, non-key frames at test will be regarded as out-of-distribution (OOD) frames. In this setup, the key frame recognition is simplified to discarding OOD frames (non-key frames) based on the quality of the BU-Net's prediction. 
We leverage three types of prediction quality to recognize key frames vs. non-key frames (OOD frames):  
i)~Interpretability provided by the prediction heatmaps, detailed in Section~\ref{sec:context}; ii)~Pixel-level epistemic uncertainty, as OOD frames are generally expected to incur a higher epistemic uncertainty, described in Section~\ref{sec:epi}; and iii)~Pixel-level aleatoric uncertainty, as the non-key frames where the object of interest is obscured or has lower visibility are expected to have a higher aleatoric uncertainty, detailed in Section~\ref{sec:alea}. Subsequently, a statistical method is used to learn to differentiate between BU-Net's uncertainty signatures on key versus non-key frames, in a calibration phase (shown in {\centering\includegraphics[width=0.025\textwidth,valign=c]{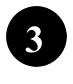}} in Fig.~\ref{fig:block_diagram}), explained in Section~\ref{sec:calibration}.
In the test, the landmark detector (BU-Net) is applied to all the frames of the video and detects landmarks on key frames based on its prediction confidence. 



\subsection{Landmark Detector}
\label{sec:backbone}
The per-frame landmark detection is formulated as a segmentation problem, where the BU-Net predicts both the heatmaps highlighting the coordinates of the landmark points, and uncertainty maps described in Sections~\ref{sec:epi} and~\ref{sec:alea}. 
The BU-Net is based on a modified U-Net architecture (network details are given in~\ref{sec:implement}). 

To generate the ground-truth masks for training, we place a circle with radius $\delta$ at the coordinates of landmarks, where $\delta$ is a hyperparameter tuned to account for the inter-observer variability in landmark labels. Here for the purpose of demonstration, we adapt a circle formation, while more complex choices of ground-truth genesis based on the underlying object could be investigated~\cite{Liu_2019_CVPR}. At test time, the center of gravity (COG) of the predicted heatmaps are used to obtain the coordinates of the landmarks. The BU-Net, only trained on key frames, generally produces valid predictions in vicinity of the key frames, and produces invalid predictions for non-key frames, where it shows an unseen event or the object of interest does not have clear visibility. Therefore, the prediction quality based on the interpretation of the heatmaps could be used as a criterion to detect the key frames discussed next. 

\subsection{Contextual Quality Control}
\label{sec:context}
The predicted heatmaps by the detector are passed through a sanity check to discard non-valid predictions, which correlate to non-key frames. Each key frame has $\tau$ landmark points, where each point is marked in the ground-truth mask with a circle of radius $\delta$. We perform a spatial sanity check on the predicted heatmap of each frame by locating all blobs with radius~$> \delta$; if $\hat \tau$ heatmaps are located in the prediction, and $\hat \tau \neq \tau$, the frame is discarded as a non-key frame.  
The BU-Net predictions tend to be invalid the further you move from the key frames, \ie the detector fails to predict enough heatmaps ($\hat \tau < \tau$) or predicts multitude of noisy heatmaps ($\hat \tau > \tau$). After this initial contextual sanity check based on the prediction heatmap, the remaining candidate key frames are assessed in terms of predictive uncertainty. The two major sources of predictive uncertainty considered are \textit{epistemic}  and \textit{aleatoric}.

 \subsection{Epistemic Uncertainty}
\label{sec:epi}
The epistemic uncertainty accounts for the ignorance in model parameters, which can be decreased by collecting more training data. 
The BU-Net, having only been trained on key frames, is expected to have a lower epistemic uncertainty on frames that have characteristics similar to the key frames, and higher uncertainty as the further you move from the key frames in time (OOD frames).
Therefore, the epistemic uncertainty of BU-Net's prediction could be used to distinguish between key and non-key frames.

Bayesian neural networks (BNNs) quantify epistemic uncertainty by learning the posterior distribution over the network's parameters ($w$). Given training frames $F$ and labels $Y$, the Bayes rule gives the posterior distribution by:


\begin{equation}
    p(w|F,Y) = \dfrac{p(Y|F,w)p(w)}{p(Y|F)}, 
\end{equation}
where the denominator is defined as:
\begin{equation}
\label{eq:integral_Bayes}
   p(Y|F) =  \int p(Y|F,w)p(w)dw.
\end{equation}

The prediction $\hat y$ for frame $\hat f$ is obtained by marginalizing over the space of possible weights:
\begin{equation}
    p(\hat y|\hat f, F, Y) = \int p(\hat y | \hat f , w, F, Y) p(w|F,Y)dw,
\end{equation}
which provides a distribution over possible outputs, and naturally gives a measure of epistemic uncertainty. 
In practice, the integral in Eq.~(\ref{eq:integral_Bayes}) is not computationally tractable and must therefore be approximated. Various approximations of BNNs exist    
that estimate the true posterior $p(w|F,Y)$, with a variational distribution $q(w)$, which is feasible to train. Popular BNN approximations include stochastic variational inference~\cite{blundell2015weight,krishnan2020improving,ovadia2019can}, multiplicative normalizing flows~\cite{louizos2017multiplicative}, stochastic batch normalization~\cite{atanov2018uncertainty}, and variational inference by Monte-Carlo (MC) dropout~\cite{gal2016dropout, kendall2017bayesian}. Also, non-Bayesian approaches generally train many deep ensemble or bootstrap models~\cite{lakshminarayanan2017simple, osband2016deep} to measure the uncertainty through predictive variance. 

In our work, U-LanD adapts the widely used MC dropout BNN approximation; 
however, the U-LanD framework is not tied to the specific choice of the uncertainty estimation method. Dropout is the regularization technique conventionally used only in the training phase of the deep models, where each node is randomly ignored (dropped) with the probability $p_{drop}$, in order to avoid over-fitting by imposing several predictive paths. Gal and Ghahramani show in~\cite{gal2015bayesian} that dropout layers give a Bernoulli-approximate variational inference of BNNs. 
Dropout is added after all the layers of BU-Net where the dropout is enabled at both training and test phases. With MC dropout enabled, each forward pass gives a sample from the approximate posterior distribution $q(w)$, obtained through a stochastic subset of the network. In the test, we run the BU-Net for $M_E$ times to use the average of the outputs from stochastic forward passes as our predicted landmark heatmap, and consider the standard deviation as a measure of epistemic uncertainty (gives $\hat u_{epi}$). 





\subsection{Aleatoric Uncertainty}
\label{sec:alea}
The aleatoric uncertainty accounts for the ambiguity in the observed data, such as the inherent noise in the measurements or  labels. 
The Bayesian landmark detector is expected to have a higher aleatoric uncertainty on noisy non-key frames, where the object of interest has low visibility. The aleatoric uncertainty of BU-Net therefore can be used to detect the key frames in time. 

Popular approaches to estimate the input-dependent aleatoric uncertainty include test-time augmentation~\cite{moshkov2020test,wang2019aleatoric}, and learning uncertainty by parameterizing the observation noise in the likelihood function~\cite{kendall2017uncertainties,le2005heteroscedastic}. In BU-Net, the input-dependent pixel-level aleatoric uncertainty is learned by modelling the observation as a sample from a Gaussian distribution. The BU-Net learns to jointly predict two output maps, namely the aleatoric uncertainty along with the landmark heatmap. 

We place a Gaussian distribution over the logits of the output layer, thus for the output pixel in row $i$ and column $j$ we have:
\begin{equation}
\label{eq:aleatoric}
\begin{aligned}
    l_{i,j} \sim \mathcal{N}(\mu_{i,j}, \sigma^{2}_{i,j}),\\
           p_{i,j} = sigmoid(l_{i,j}),
\end{aligned}
\end{equation}
where $l$ and $p$ denote the output logit and the landmark probability for each pixel, respectively, and $\mu$ and $\sigma$ are parameters learned by the network, where $\sigma$ correlates to the pixel-level aleatoric uncertainty. The reparametrization trick is used with Eq.~(\ref{eq:aleatoric}) to enable learning by backpropagation:
\begin{equation}
\label{eq:rep_trick}
    l_{i,j}  = \mu_{i,j} + \sigma_{i,j} \epsilon. \quad \quad \epsilon \sim \mathcal{N}(0,1) .
\end{equation}

 The network is trained with two loss functions, namely, soft Dice and the weighted binary cross-entropy (WBCE), comparing the predicted heatmap ($p$ in Eq.~(\ref{eq:aleatoric})) to the ground-truth landmark masks. The weights of the WBCE are assigned inversely proportionate to the number of pixels in landmark versus background, normalizing the effect of imbalanced classes. 
 While there is no analytic solution to integrate out the Gaussian distribution in Eq.~(\ref{eq:rep_trick}), it can be approximated by Monte-Carlo integration using $M_A$ samples at training:
\begin{equation}
    p_{i,j} = \dfrac{1}{M_A}\sum_{i=1}^{M_A} sigmoid(\mu_{i,j} + \sigma_{i,j} \epsilon), \quad \quad \epsilon \sim \mathcal{N}(0,1) ,
\end{equation}
whereas no sampling is needed at test ($\epsilon=0$), and the network learns to predict $\mu$ (which gives the landmark prediction heatmap) and $\sigma$ (which gives a measure of aleatoric uncertainty, \ie gives $\hat u_{alea}$) for each input frame in one forward pass. Here for the purpose of demonstration, we assume a simple Gaussian distribution for the per-pixel labels; however, the U-LanD framework does not have a specific assumption about the choice of the method for aleatoric uncertainty estimation. 

\subsection{Statistical Calibration}
\label{sec:calibration}
BU-Net provides three criteria of prediction confidence, namely contextual quality control, pixel-level epistemic uncertainty, and pixel-level aleatoric uncertainty. At test time, the indices of the key frames are unknown, and the above three confidence criteria are used to detect key frames vs. non-key frames, as shown in {\centering\includegraphics[width=0.025\textwidth,valign=c]{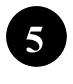}} in Fig.~\ref{fig:block_diagram}. If a frame passes the three confidence criteria, it will be added to a pool of key frame measurements (see {\centering\includegraphics[width=0.025\textwidth,valign=c]{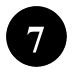}} in Fig.~\ref{fig:block_diagram}), otherwise, it will be discarded as a non-key frame (see {\centering\includegraphics[width=0.025\textwidth,valign=c]{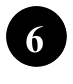}} in Fig.~\ref{fig:block_diagram}). The first criterion (contextual quality control) is a spatial sanity check based on the prediction heatmaps, explained in Section~\ref{sec:context}, which does not require a calibration phase. After passing this first criterion, the predictions are evaluated in terms of epistemic (Section~\ref{sec:epi}) and aleatoric (Section~\ref{sec:alea}) uncertainty, for which a threshold of validity needs to be tuned in a calibration phase. 
We set aside a part of unseen training data, called \textit{validation-calibration} set $\X_{calib}$ with $N^'$ samples. After BU-Net's training is done, in a calibration phase, we run BU-Net on the key frames of $\mathcal{X}_{calib}$ and find the $sum()$ of pixel-level uncertainty values for each prediction, \ie a vector of size $N^'$ is obtained, for both the aleatoric and epistemic maps. Thereafter, the mean ($\mu_{alea}$ and $\mu_{epi}$) and standard deviation ($\sigma_{alea}$ and $\sigma_{epi}$) of uncertainty vectors are calculated. In test time, we measure $Z_{score} = {\lvert sum(\hat u) - \mu \rvert}/{\sigma}$ for both the aleatoric and epistemic uncertainty maps ($\hat u_{epi}$ and $\hat u_{alea}$) of each frame, and discard the frame as a non-key frame (OOD frame) if $Z_{score} > \xi$. Further, a temporal sanity check is used, where each frame is only added to the pool of key frames if a temporal window of $\lambda$ frames in its adjacency have also passed the above confidence criteria. 

Using the above confidence criteria, U-LanD automatically selects the key frames for landmark measurement directly based on the quality of the predicted landmarks by BU-Net. We use the pool of detected key frames to obtain the key landmark measurements.  The experiment details and the evaluations showing the effectiveness of this approach are presented in the following Section.

\begin{figure*}[t]
\centering
\small
\begin{tabular}{cccc}
a) Sample Non-key Frame  & \hskip -0.06in & \hskip -0.12in Epistemic &  \hskip -0.12in Aleatoric  \\
[-0.1in]
\subfloat{\includegraphics[width=0.22\textwidth,valign=c]{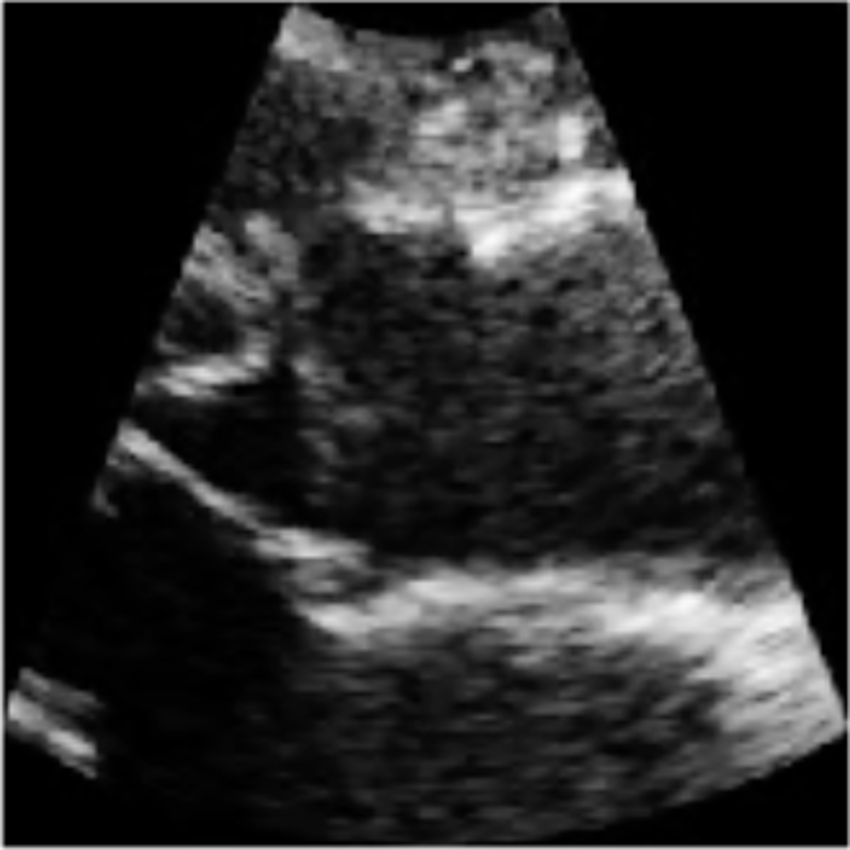}} &
\hskip -0.06in
\subfloat{\includegraphics[width=0.018\textwidth,valign=c]{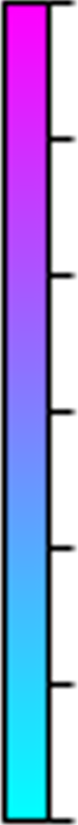}} &
\hskip -0.12in
\subfloat{\includegraphics[width=0.22\textwidth,valign=c]{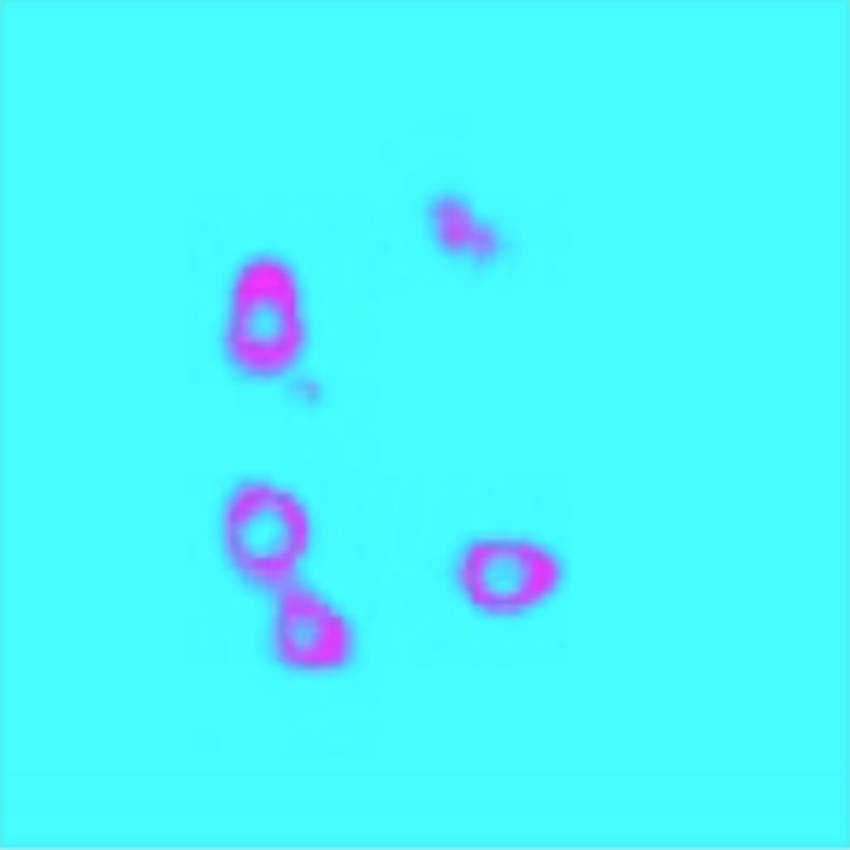}} &
\hskip -0.12in
\subfloat{\includegraphics[width=0.22\textwidth,valign=c]{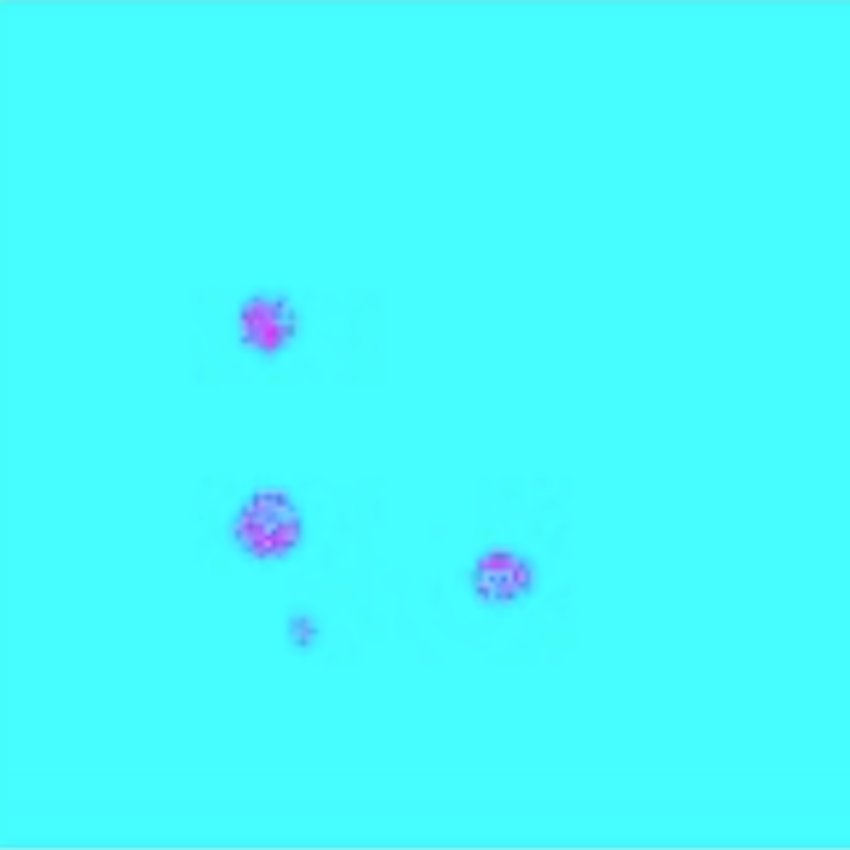}} \\
[0.7in] \\
b) Sample Key Frame & \hskip -0.06in & \hskip -0.12in Epistemic & \hskip -0.12in Aleatoric \\ 
[-0.1in]
\subfloat{\includegraphics[width=0.22\textwidth,valign=c]{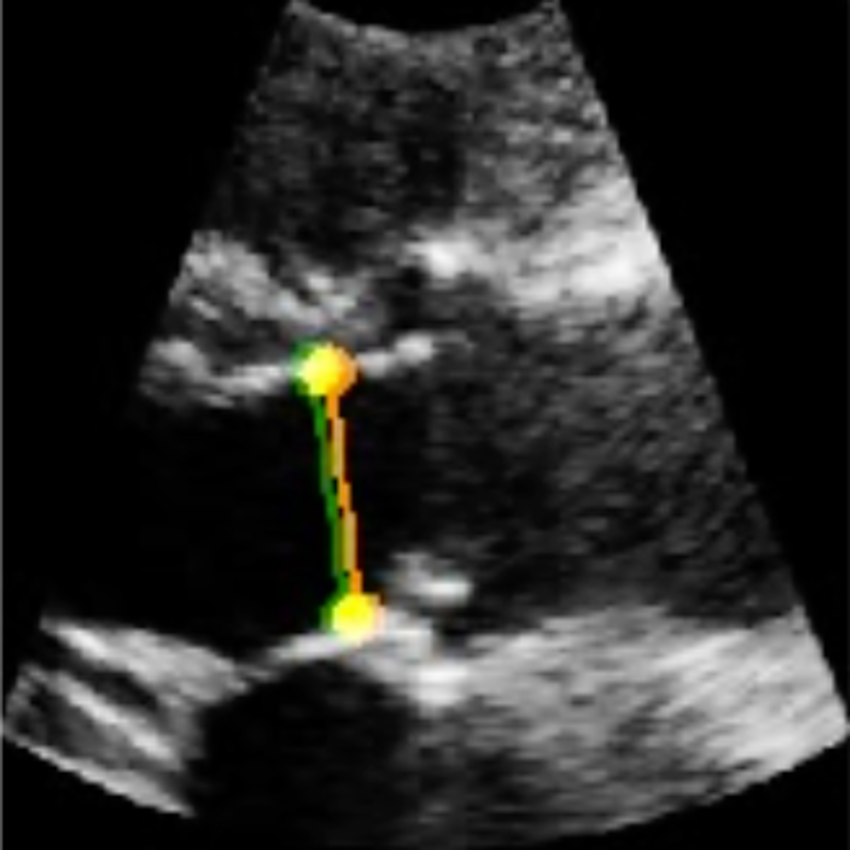}} &
\hskip -0.06in
\subfloat{\includegraphics[width=0.018\textwidth,valign=c]{Figures/Fig-Visual/bar.png}} &
\hskip -0.12in
\subfloat{\includegraphics[width=0.22\textwidth,valign=c]{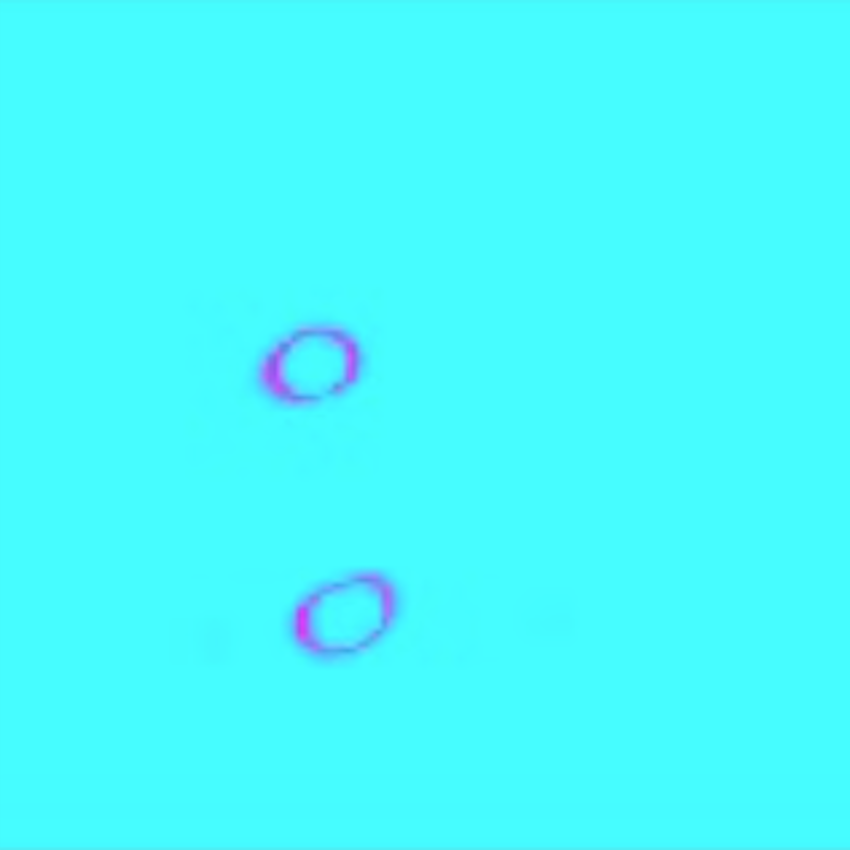}} &
\hskip -0.12in
\subfloat{\includegraphics[width=0.22\textwidth,valign=c]{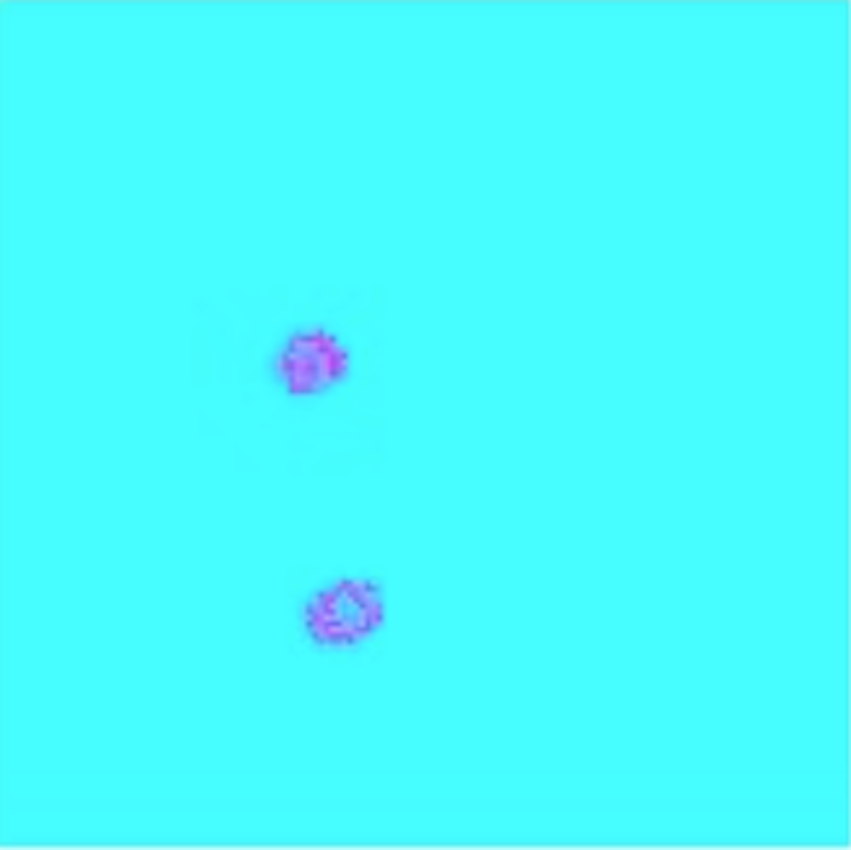}} 
\\

\end{tabular}

    \caption{Sample frames of a test video, along with their corresponding aleatoric and epistemic uncertainty maps. a) Sample non-key frame, automatically discarded due to high uncertainty (no landmark prediction). b) Sample key frame with predicted landmarks superimposed on the frame. The targeted object is LVOT in heart ultrasound videos; the landmarks are used to measure LVOT length (the line between the two dots). Image (b) shows the predicted landmarks (green), ground-truth landmarks (orange), and their overlap (yellow). The prediction error (mean absolute difference in length) for this case is 0.18 mm.  Further, sample video results are presented in the supplementary materials.}
    \label{fig:visual-res}
\end{figure*}



\begin{figure*}[t!]
    \centering
    \begin{tabular}{ccc}
        \includegraphics[width=0.29\textwidth]{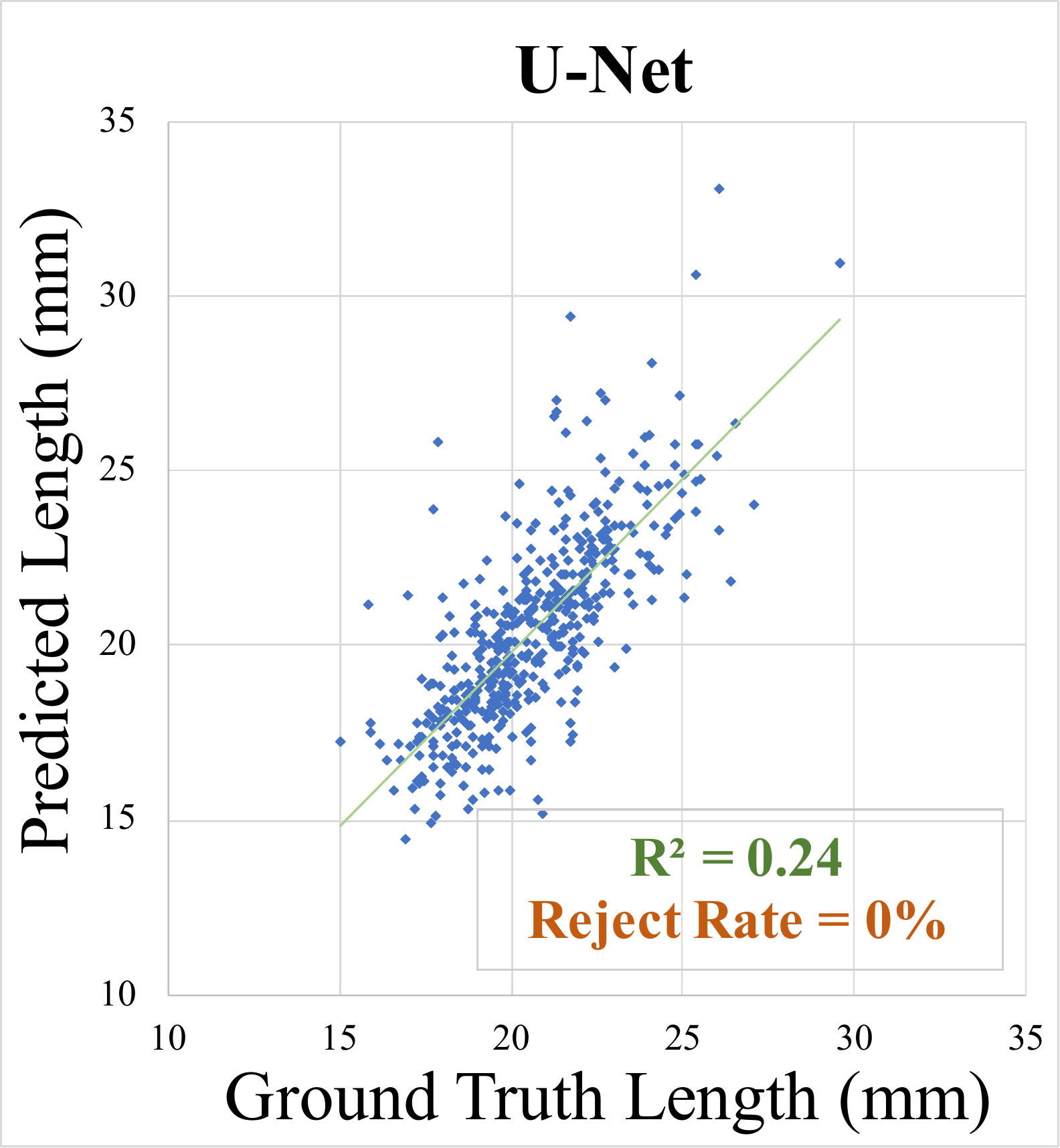}
        & \includegraphics[width=0.29\textwidth]{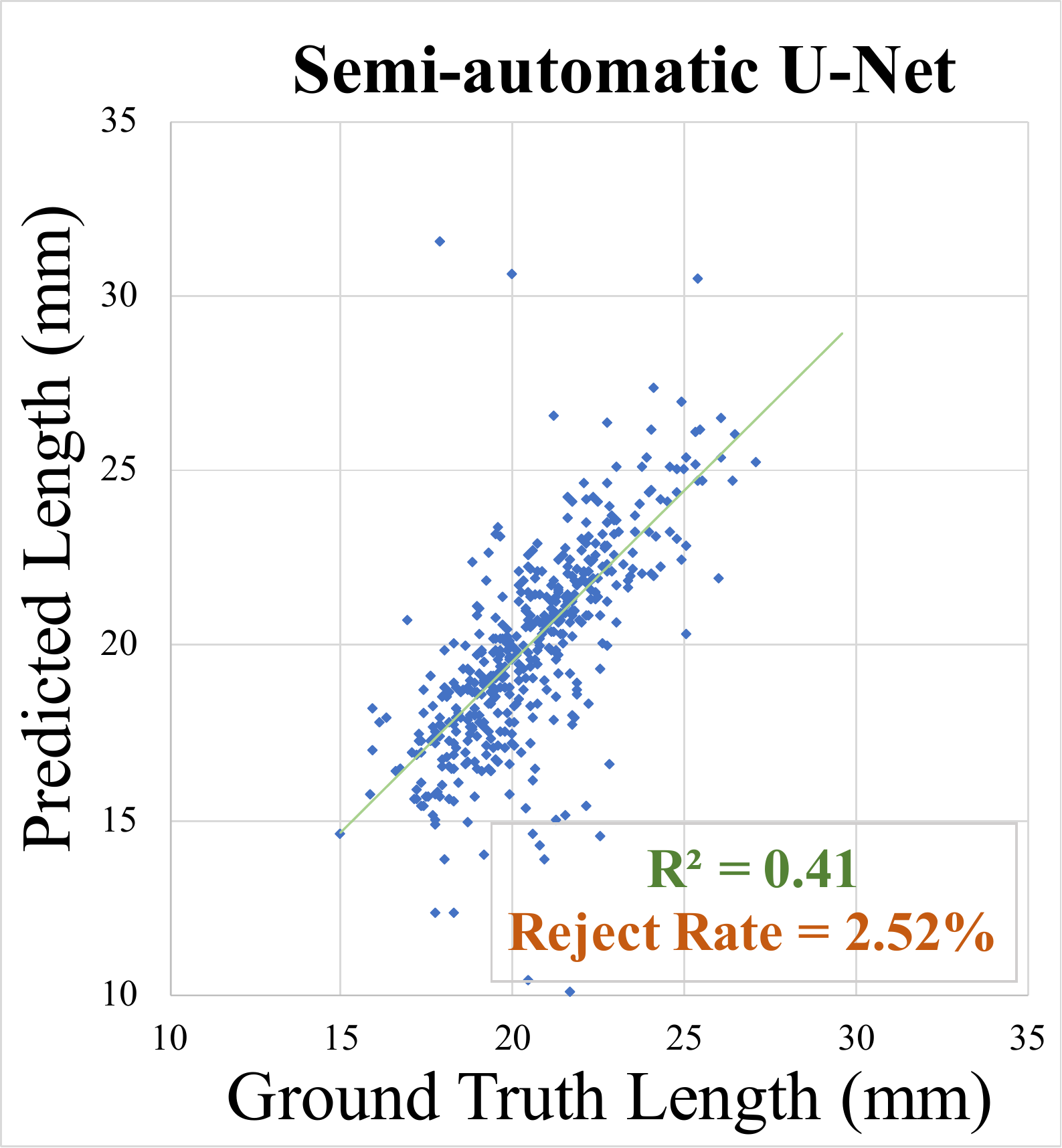}
        & \includegraphics[width=0.29\textwidth]{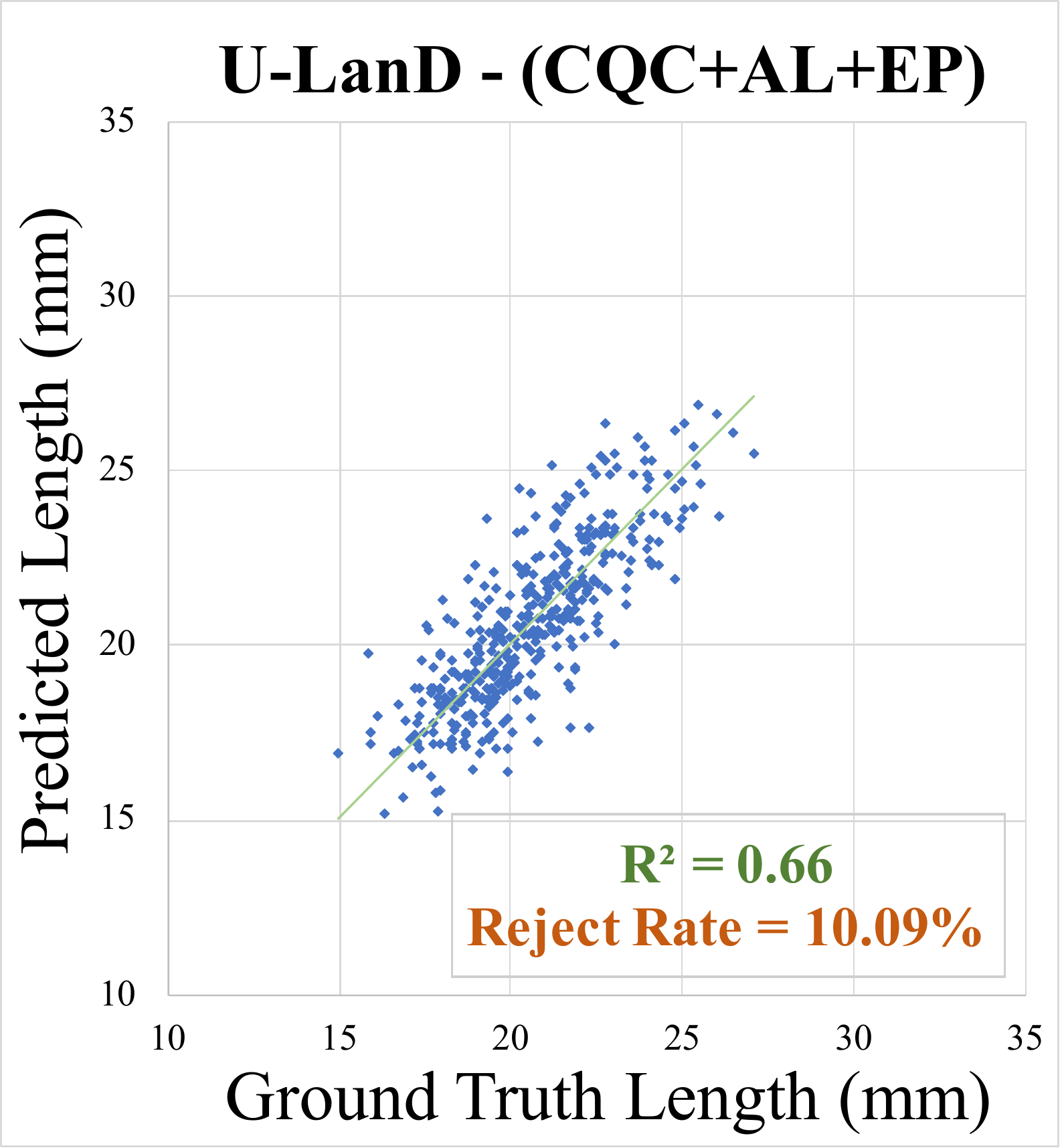}
        \\
    \end{tabular}
    \caption{Object length measured using predicted landmarks versus ground-truth length. Left: U-Net analyzing all the frames. Middle: Semi-automatic approach, where video key frames are selected by an expert human. Right: U-LanD (Proposed). The targeted object is LVOT in heart ultrasound videos, where LVOT length is measured using the predicted landmarks.} 
    \label{fig:compare}
\end{figure*}


\begin{table*}[t!]
\caption{Evaluation of the video landmark measurement performance. Best results are in bold.}
\centering
\resizebox{1.0\textwidth}{!}{%
    \begin{tabular}{ c | c | c | c | c | c | c}
    Method &  \makecell{$R^2$ \\ (\%)} & \makecell{$\Delta R^2$\\ (\%)} & \makecell{MAE\\ (mm)}  & \makecell{STD\\ (mm)} & \makecell{MAX \\ (mm)} & \makecell{Reject \\ Rate ($\%$)} \\   \hline
    I) U-Net; per-frame analysis of all frames~\cite{karpathy2014large,liu2020efficient,perazzi2017learning} & 24 &  n/a & 1.73 & 3.14	& 42.81	& 0 \\ 
    II) Semi-automatic U-Net; analyzing the key frames~\cite{girdhar2019video,li2020spatio,mishra2020real}
    &  41 & +71 & 1.50 & 2.14 & 25.39	& 2.52 \\
     U-LanD (CQC+AL+EP) - \textbf{Proposed} &  \textbf{66} & \textbf{+175} & \textbf{1.08} & \textbf{0.89} & \textbf{4.66}	& 10.69 \\
    \end{tabular}}
\label{table:res}
\end{table*}

\begin{figure*}[t!]
    \centering
    \begin{tabular}{cc}
        \includegraphics[width=0.29\textwidth]{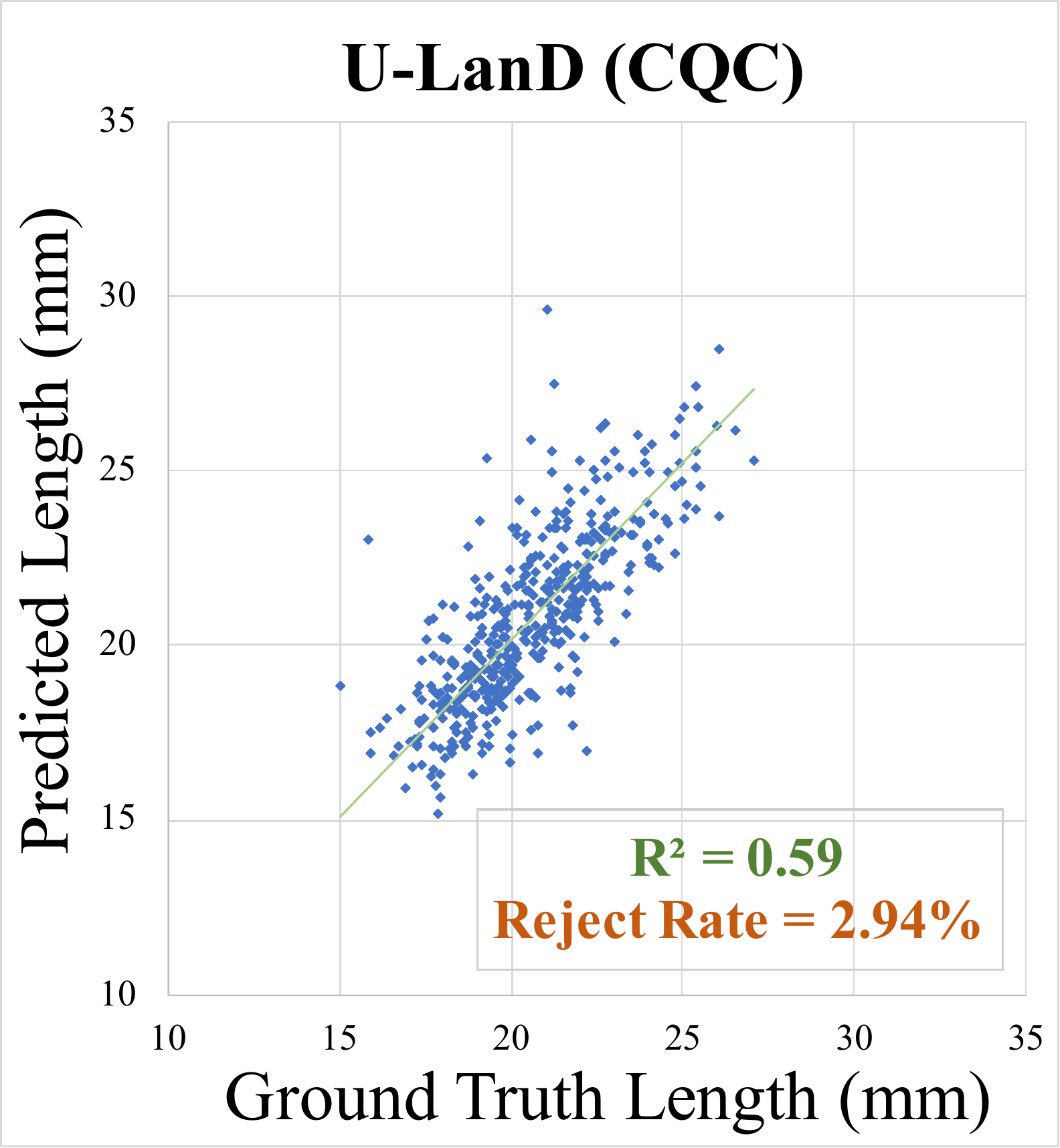} &      \includegraphics[width=0.29\textwidth]{Figures/Fig-Graphs/uland-cqc-al-ep.pdf}
        \\
    \end{tabular}
    \caption{Object length measured using predicted landmarks versus ground-truth length. Left: U-LanD using contextual quality control (CQC). Right: U-LanD using CQC and predictive uncertainty; aleatoric (AL) and epistemic (EP). The targeted object is LVOT in heart ultrasound videos, where LVOT length is measured using the predicted landmarks.}
    
    \label{fig:ablation}
\end{figure*}

\section{Experiments}
\subsection{Dataset}
The data used for the experiments are collected from Picture Archiving and Communication System at 
Vancouver General Hospital (VGH), Canada, 
following approval from the Medical Research Ethics Board in coordination with the privacy office. 
We gathered echo videos of 4,493 patients, captured by a variety of ultrasound devices, namely Philips iE33, Vivid~i/7/9/95, Sonosite, and Sequoia. The echo cines show the parasternal long axis (PLAX) view of the heart. For all the echo videos, the ground-truth locations of $\tau=2$ points regarding LVOT landmarks are only available for one frame throughout the video. The ground-truth annotations are derived from the archived information of the patients, originally provided by expert cardiologists. The videos are split into train and test set with a ratio of 90\% and 10\%, respectively. Also, 10\% of training data are set aside as the $\X_{calib}$ to learn the  statistics of BU-Net uncertainty signatures, as explained in Section~\ref{sec:calibration}. There is no overlap between patients in the train, validation-calibration, and test set.

For echo LVOT landmark detection in clinical settings, the expert cardiologist examines the echo video, selects a key frame around mid-systole phase of the heart, where the LVOT landmarks are clearly visible, and annotates two points used to measure LVOT length. Sample echo frames and the corresponding visual results and uncertainty maps are presented in Fig.~\ref{fig:visual-res}.
In U-LanD, the BU-Net takes all the frames of the video and predicts landmarks for the recognized key frames, \ie the frames where the prediction confidence passes the quality criteria explained in Section~\ref{sec:calibration}, and otherwise discards low confidence predictions regarding them as non-key frames. 

Due to clinical diagnostic needs, the expert cardiologists usually tend to annotate the LVOT landmarks at the key frame that shows the maximum lengths of the LVOT. In U-LanD, if the predicted LVOT landmarks from the key frames were perfectly correct, we could just select the maximum length among the key frames to be reported as our predicted LVOT length; however, we have a pool of predicted key frame measurements, and selecting the single maximum value in the pool might be prone to error, we therefore instead choose to select the 75\% percentile of the predicted lengths to be reported as the LVOT length. This approach helps all the high quality predicted landmarks in key frames contribute to the reported measurement, and improve the accuracy of the predicted LVOT length by further cancelling out the noisy observations. 

    
    

\begin{table*}[ht]
\caption{Ablation studies. Best results are in bold.}
\centering
\resizebox{0.7\textwidth}{!}{%
    \begin{tabular}{ c | c | c | c | c | c | c}
    Method &  \makecell{$R^2$ \\ (\%)} & \makecell{$\Delta R^2$\\ (\%)} & \makecell{MAE\\ (mm)}  & \makecell{STD\\ (mm)} & \makecell{MAX \\ (mm)} & \makecell{Reject \\ Rate ($\%$)} \\   \hline
    U-LanD (CQC) &  59 & +146 & 1.19 & 1.09 & 8.57	& 2.94 \\ 
    U-LanD (CQC+AL)&  63 & +162 & 1.12	& 0.94 & 5.15	& 5.24 \\
    U-LanD (CQC+EP)&  63 & +162 & 1.11 & 0.94	& 6.78 	& 9.85 \\
     U-LanD (CQC+AL+EP)&  \textbf{66} & \textbf{+175} & \textbf{1.08} & \textbf{0.89} & \textbf{4.66}	& 10.69 \\
    \end{tabular}}
\label{table:ablation}
\end{table*}

\subsection{Implementation Details}
\label{sec:implement}
The landmark detector BU-Net is based on a U-Net architecture with three down-sampling blocks and three reverse up-sampling blocks, where skip connections concatenate feature maps from the encoder to the corresponding layer in the decoder path. Each down-sampling block includes two convolutional layers with kernel size $3\times3$ and a maxpooling of size $2\times2$. The decoder mirrors the down-sampling blocks, except transpose convolution is used for upsampling. The number of filters in the first convolutional layer is $32$ which is doubled after each down-sampling step. The covolutional layers are followed by batch normalization ($\textrm{momentum}=0.8$), ReLU, and dropout ($p_{drop}=0.2$). The last layer includes two output maps (as Section~\ref{sec:alea}), namely predicted landmark heatmap ($1\times1$ convolutional kernel followed by sigmoid) and aleatoric uncertainty map ($1\times1$ convolutional kernel followed by softplus). The number of MC integral samples at training time used for aleatoric uncertainty learning is $M_A=100$. MC dropout samples in test are $M_E=30$. BU-Net is run once with dropout disabled to obtain aleatoric uncertainty in one pass. The input images are resized to $224\times224$ pixels. The radius of the circles placed over the location of ground-truth landmark points in training masks is $\delta=7$, the OOD threshold is $\xi =1$, and the size of the temporal window is $\lambda=5$, which are hyperparameters tuned using a validation set. We use the Adam optimizer with $\beta_1=0.9$, $\beta_2=0.999$, and a learning rate initially set to $1e-3$. On the fly data augmentation is done in each iteration, including random shift, rotation, zoom, and gamma transform.


\subsection{Evaluations}
We compare the predicted LVOT length by U-LanD against the ground-truth LVOT length in the patient's record based on the cardiologist's examination of the echo video. We calculate $R^2$ score, the square of the Pearson product-moment correlation coefficient; $R^2 \in [0\%,100\%]$, where 0\% denotes no correlation, and 100\% denotes total correlation. We also report the percentage of improvement in $R^2$ over the baseline method ($\Delta R$), mean absolute error (MAE), standard deviation (STD) of absolute error, and worst-case performance, \ie maximum absolute error (MAX). 
Table~\ref{table:res} presents the comparison with two state-of-the-art paradigms for video object detection. 

\textbf{Paradigm I)} All the frames of the videos are analyzed for object measurement~\cite{karpathy2014large,liu2020efficient,perazzi2017learning}. To have a comparative evaluation of U-LanD against paradigm I, the non-Bayesian U-Net, as a state-of-the-art base model, is used to predict landmarks on all the frames. 
The network architecture and the method used to calculate LVOT length are the same in the applied U-Net and U-LanD, except there are no measures of prediction confidence in the baseline U-Net to be used to discard uncertain (non-key frame) landmarks through time. The results of this experiment can be found in the first line of Table~\ref{table:res}, and the scatter plots comparing predicted LVOT length to ground-truth are given in Fig.~\ref{fig:compare}. 

\textbf{Paradigm II)} 
Solving the key frame video object detection in two steps, II-A) a temporal model (\eg attention RNN, 3D/(2+1)D CNN, transformer)~\cite{girdhar2019video,li2020spatio,mishra2020real,tran2018closer,yan2018deep} is trained to detect the indices of the key frames, II-B) followed by object detection at the recognized key frames. In order to compare U-LanD framework against paradigm II, we consider a semi-automatic approach, where the ground-truth indices of the key frames are suggested by the cardiologist, followed by landmark detection by U-Net on these ground-truth key frames (called \textit{semi-automatic U-Net}); the semi-automatic U-Net could be an ideal case of paradigm II, where there is no error in step II-A, by using the ground-truth key frame index. 
The semi-automatic U-Net results can be found in the second line of Table~\ref{table:res}, and the scatter plots are presented in Fig.~\ref{fig:compare}.

As can be seen in Table~\ref{table:res} and Fig~\ref{fig:compare}, U-LanD significantly improves the state-of-the-art  non-Bayesian base model. U-LanD's superiority could be attributed to its ability to analyze intervals of the video specifically selected by the Bayesian landmark detector (BU-Net) when the prediction confidence is high enough for accurate landmark measurement, particularly discarding the noisy frames in the challenging echo videos. Futhermore, the right-most column of Table~\ref{table:res} and the orange text on Fig.~\ref{fig:compare} and Fig.~\ref{fig:ablation} (Reject Rate) shows the percentage of rejected videos by U-LanD, when the model has automatically detected that the video quality is not good enough for accurate landmark measurement. The rejected rate in semi-automatic U-Net denotes videos where the U-Net has failed to predict two landmarks needed to calculate LVOT distance on the ground-truth key frame index. 

\textbf{Ablation Studies:} The ablation study is given in Fig.~\ref{fig:ablation} and Table~\ref{table:ablation}; U-LanD (CQC): only using the contextual quality control (Section ~\ref{sec:context}), U-LanD (CQC+AL): CQC and aleatoric uncertainty (Section~\ref{sec:alea}), U-LanD (CQC+EP): CQC and epistemic uncertainty (Section~\ref{sec:epi}), and U-LanD (CQC+AL+EP): proposed method using all three criteria of prediction confidence. As can be seen, the three criteria of quality control, namely CQC, AL, and EP have complementary effects, significantly improving the results by accounting for different aspects of prediction confidence.  

\textbf{Further Analysis:}
Compared to non-Bayesian base, U-LanD has $M_E$ runtime overhead for epistemic uncertainty sampling; 
however, for time-sensitive applications, the proposed U-LanD (CQC+AL) could be the suitable choice, where the uncertainty is obtained in one forward pass (no need for sampling), while the $R^2$ score of CQC+AL is decreased only by 3\% compared to the U-LanD's full mode. Compared to 3D or attention-based models conventionally used for video segmentation and object measurement, U-LanD is built upon a small-size 2D (per-frame) model, which enables its integration into memory-limited embedded or mobile systems. U-LanD almost has no memory overhead compared to the non-Bayesian counterpart (less than 1e-3\%); only one extra feature map is added in the last layer of the detector to predict aleatoric uncertainty.

\section{Conclusions}
In this paper we proposed U-LanD, a Bayesian framework for joint key frame and landmark detection in videos with extremely sparse and noisy labels. U-LanD leveraged the variations of uncertainty through time as an unsupervised signal to detect landmarks on key frames of the video. We demonstrated U-LanD on a challenging dataset of heart ultrasound series, where only one frame in each training video is annotated, and yet the annotations are noisy clinical labels. For experiments, we gathered a large-scale video dataset of echo series from 4,493 patients, showing the U-LanD efficiency; improving the $R^2$ score of state-of-the-art non-Bayesian counterpart with a noticeable margin of 42\% (+175\%), with  no extra-label cost, and almost no memory overhead. U-LanD (fully automatic) can also outperform the semi-automatic video landmark detection by 25\% (+61\%) in $R^2$ score. 
U-LanD can achieve good results even with simple choices of backbone detector and OOD rejection method. Future work could include investigating temporal uncertainty in multi-object detection and tracking.

\bibliography{refs.bib} 
\bibliographystyle{IEEEtran}

\end{document}